\documentclass[10pt,journal,cspaper,compsoc]{IEEEtran}
\usepackage{graphicx}
\usepackage{latexsym}
\newtheorem{df}{Definition}
\newtheorem{tm}{Theorem}
\newtheorem{lm}{Lemma}
\newtheorem{pr}{Proposition}
\newtheorem{am}{Assumption}
\newcommand{\bsquare}{\hbox{\rule{6pt}{6pt}}}

\ifCLASSOPTIONcompsoc
\else
\fi
\ifCLASSINFOpdf
\else
\fi
\begin{document}
%
\title{Causal Discovery in a Binary Exclusive-or Skew Acyclic Model: BExSAM}
%
%
%
%

\author{Takanori~Inazumi, 
        Takashi~Washio,~\IEEEmembership{} 
        Shohei~Shimizu, Joe~Suzuki,~\IEEEmembership{} 
        Akihiro~Yamamoto, 
        and Yoshinobu~Kawahara,~\IEEEmembership{}

\IEEEcompsocitemizethanks{\IEEEcompsocthanksitem T. Inazumi, T. Washio, 
              S. Shimizu and Y. Kawahara are with the Institute of Scientific 
              and Industrial Research, Osaka University, 8-1 Mihogaoka, 
              Ibaraki, Osaka, Japan.\protect\\
E-mail: washio@ar.sanken.osaka-u.ac.jp
\IEEEcompsocthanksitem J. Suzuki is with Graduate School of Science, 
Osaka University, and A. Yamamoto is with Graduate School of Informatics, 
Kyoto University.
\IEEEcompsocthanksitem A preliminary result has been presented in an 
international conference: Discovering causal structures in binary exclusive-or 
skew acyclic models, Proc. the 27th Conference on Uncertainty in Artificial 
Intelligence (UAI2011), pp.373-382, 2011.}
\thanks{}}

%
%

\markboth{}%
{Inazumi \MakeLowercase{\textit{et al.}}: Causal Discovery in a Binary Exclusive-or Skew Acyclic Model: BExSAM}
%



\IEEEcompsoctitleabstractindextext{%
\begin{abstract}
Discovering causal relations among observed variables in a 
given data set is a major objective in studies of statistics 
and artificial intelligence. Recently, some techniques to 
discover a unique causal model have been explored based 
on non-Gaussianity of the observed data distribution. However, 
most of these are limited to continuous data. In this paper, 
we present a novel causal model for binary data and propose 
an efficient new approach to deriving the unique causal model 
governing a given binary data set under skew distributions of 
external binary noises. Experimental evaluation shows excellent 
performance for both artificial and real world data sets.
\end{abstract}

}

\maketitle

\IEEEdisplaynotcompsoctitleabstractindextext

%
\IEEEpeerreviewmaketitle

\section{Introduction}

\IEEEPARstart{M}{any} approaches to causal inference and learning 
of Bayesian network structures have been studied in statistics and 
artificial intelligence~\cite{Pearl:2000, Spirtes:2000}. 
Most of these derive candidate causal models from an observed 
data set by assuming acyclicity of the causal dependencies. 
These mainly use information from second-order statistics 
such as correlations of the observed variables, and narrow 
down the candidate directed acyclic graphs (DAGs) by using 
some constraints and/or scoring functions. However, these 
approaches often produce multiple candidate causal structures 
for the data set because of the Markov equivalence or the 
local optimality of the structures.

Recently, some non-Gaussianity-based approaches applied 
to a linear acyclic model (LiNGAM) have been 
proposed~\cite{Shimizu:2006, Shimizu:2009, Shimizu:2011}. 
Such approaches derive a unique causal order of observed variables 
in a linear acyclic structural equation model under the condition that 
external noises driving the variables are non-Gaussian and jointly 
independent, and estimate a unique model based on the derived 
causal order. However, these approaches require linearity of the 
objective system. Recent studies extended this principle to non-linear 
models~\cite{Hoyer:2009, Zhang:2009}. They clarified conditions to 
identify unique causal orders in bivariate non-linear and post 
non-linear (PNL) models, and applied these conditions to derive 
candidate causal orders in their multivariate models and estimate 
models based on these orders. However, their 
applicability is limited to the model consisting of continuous 
variables and smooth non-linear functions, and identifying 
a unique causal order in the multivariate model is not guaranteed. 
More recent work further extended the principle for a bivariate model 
where the two variables have ordered discrete values~\cite{Peters:2011a}. 
However, this did not address the identification of a unique causal 
order in a multivariate model and the estimation of the model 
under the identified order.

In contrast, many real world domains such as computer networks, 
medicine~\cite{Pearl:2000}, bioinformatics~\cite{Veflingstada:2007, Shi:2007} 
and sociology~\cite{Spirtes:2000}, maintain recently-accumulated 
stochastic binary data sets, and practitioners need to discover 
the causal structures and structural models governing the data 
sets for various purposes. However, to the best of our knowledge, 
no past studies have addressed principles and algorithms to practically 
derive a unique causal order and a causal model following the 
order for a given binary data set. In this regard, our objective 
in this paper is to propose a novel and practical approach to 
discovering such a causal order and the associated model within 
a given stochastic binary data set under some feasible assumptions. 

In the next section, we briefly review some related work to indicate 
important technical issues. In the third section, we introduce a 
novel binary exclusive-or (EXOR) skew acyclic model, termed 
``{\it BExSAM},'' to represent an objective system, and characterize 
the model with respect to causal order identification and causal model 
estimation. In the fourth section, we propose novel criteria and 
algorithms for causal order identification and causal model 
estimation based on the model characterization. In the fifth section, 
we present an experimental evaluation of this approach using both 
artificial and real world data sets.

\section{Related Work}

Many studies on causal inference in statistics and learning of 
Bayesian network structures have concentrated on developing principles 
for efficiently focusing on candidate causal structures of a given 
data set within a feasible search space by using information from 
second-order statistics of the data set~\cite{Pearl:2000, Spirtes:2000}. 
This has arisen because 
exhaustive searches are intractable as the number of possible directed 
acyclic graphs (DAGs) grows exponentially with the number of variables. 
To address this issue, constraint-based approaches such as the PC and 
the CPC algorithms~\cite{Spirtes:2000, Ramsey:2006} 
and score-based approaches such as the GES 
algorithm~\cite{Chickering:2002} 
have been studied 
for both continuous and discrete variables. However, these admit 
multiple solutions because of the Markov equivalence and the 
local optimality of those solutions in many cases, and thus often 
fail to generate a uniquely identifiable 
causal structure. They also need the assumption of faithfulness, 
implying that correlations between the variables are entailed by the 
graph structure only.

A recent technique LiNGAM~\cite{Shimizu:2006} formulates the causal 
DAG structure search and the structural modeling in the form of an 
independent component analysis (ICA), that ensures the existence of 
a unique global optimum under assumptions of linear relations among 
the observed variables, non-Gaussianity and joint independence of 
their external noises. Such a condition that enables the identification 
of a unique causal order of observed variables in a model, even if some 
other structural models are Markov equivalent to the model, is called 
an ``{\it identifiability condition}.'' However, this may often provide 
a locally optimal solution through the nature of its greedy search. 
In contrast, the more recent DirectLiNGAM~\cite{Shimizu:2009, Shimizu:2011} 
efficiently derives a uniquely identifiable solution under the same 
identifiability condition through its iterative search for 
exogenous variables, {\it i.e.}, causally top upstream variables, 
by applying simple bivariate linear regressions and independence 
measures. The other notable advantage of these LiNGAM approaches is 
that they do not need the faithfulness assumption. Two 
studies~\cite{Hoyer:2009, Zhang:2009} proposed extensions of 
the principles of LiNGAM to a non-linear additive noise model and 
a post-nonlinear (PNL) model respectively. However, these two 
studies presented the identifiability conditions for two variable 
cases only, {\it i.e.}, ``{\it bivariate identifiability 
conditions},'' and did not provide ``{\it multivariate identifiability 
conditions}'' or the algorithms to identify unique identifiable 
causal orders in multivariate models as mentioned in the former 
section. Another study~\cite{Mooij:2009} proposed a novel 
regression to allow causal inference in a non-linear additive 
noise model containing multiple variables by introducing HSIC 
(Hilbert-Schmidt independence criterion). However, the 
identifiability of a unique solution is not ensured because of 
the non-convexity of the regression problem. 

In contrast to these studies for continuous variables, only a few 
studies have addressed the issue of discovering causal structure for 
discrete variable sets based on particular characteristics 
of their data distributions. A study on this topic was recently 
reported in~\cite{Peters:2011a}. It assessed the identifiability 
of a unique causal order and an algorithm to find a model 
entailed by the order for integer variables in a finite range 
and/or a cyclic range having a modulus. However, the focus is on 
bivariate models and their bivariate identifiability conditions only. 
Another study~\cite{Sun:2007} proposed a principle for finding a 
causal order of binary variables to explain a given sample 
distribution by mutually independent conditional probability 
distributions named Markov kernels. However, its applicability 
is limited to very simple Boolean relations because of the high 
complexity of the kernel functions for the generic cases. 

More recent work showed that the acyclic causal structure in a 
multivariate model is identifiable if the causal relation on every 
pair of variables conditioned by all other variables in the model 
is bivariate identifiable~\cite{Peters:2011b}. It also indicated 
that the faithfulness assumption is not needed for the bivariate 
identifiability based approach to deriving a unique identifiable 
causal order from a given data set. Furthermore, the multivariate  
identifiability of all aforementioned models under their respective 
bivariate identifiability conditions was shown. A generic algorithm 
for deriving candidate identifiable causal structures from given 
data sets based on these results was also demonstrated. However, 
the algorithm and the independence measure used in it are not 
adapted to the model consisting of mutivariate discrete variables, 
and the applicability of the algorithm was confirmed only for 
linear and non-linear additive noise models containing up to four 
continuous variables. 

On the other hand, many real world applications need to discover 
a feasible causal order and a causal model entailed by the 
order from a given binary data set. Because binary variables 
do not constitute a continuous algebra, we need to develop a 
structural model of acyclic relations among the binary variables 
for other algebraic systems such as a Boolean algebra. In addition, 
we need to apply a binary data distribution such as a Bernoulli 
distribution instead of Gaussian or non-Gaussian distributions in 
modeling stochastic characteristics of the variables, and to use 
adapted measures to evaluate the independence of the binary 
distribution. Moreover, we have to design novel algorithms for 
both causal order identification and model estimation under 
the order based upon characteristics of the structural 
model and the data distribution. In the following sections, 
we present our ideas concerning these issues.

\section{Proposed Model}

\subsection{BExSAM}

We first introduce a novel structural model representing 
generic acyclic causal relations among binary variables. 
\begin{df}\label{dfbexsam}
Given $d \geq 1$, let $e_k \in \{0,1\}$ for all $k=1,\dots,d$ 
be jointly independent random variables, 
$f_k:\{0,1\}^{k-1} \rightarrow \{0,1\}$ deterministic Boolean 
functions and 
\[x_k = f_k(x_1,\dots,x_{k-1}) \oplus e_k,\]
where $f_1$ is constant, and $\oplus$ denotes 
the EXOR operation defined by Table~1.\hfill $\bsquare$
\end{df}
Every external noise $e_k$ affects its corresponding 
variable $x_k$ via an EXOR operation. Each $f_k$ expresses any 
deterministic binary relation without loss of generality 
because such a relation is always represented by Boolean 
algebraic formulae~\cite{Gregg:1998}. 
We further assume a skew Bernoulli distribution of every 
noise $e_k$ as follows.
\begin{am}\label{amskew}
We assume that the probability $p_k$ of $e_k=1$ satisfies 
$0 < p_k < 0.5$ for all $k=1,\dots,d$.\hfill $\bsquare$
\end{am}
This assumption covers the case $0.5 < p_k < 1$ without 
loss of generality, becasue $x_k = f_k(x_1,\dots,x_{k-1}) \oplus e_k$ 
is equivalent to $x_k = \bar{f}_k(x_1,\dots,x_{k-1}) \oplus \bar{e}_k$ 
where the probability $\bar{p}_k$ of $\bar{e}_k=1$ is $1-p_k$ satisfies 
$0 < \bar{p}_k < 0.5$\footnote{$\bar{f}_i$ and 
$\bar{e}_i$ are $f_i \oplus 1$ and $e_i \oplus 1$, respectively.}. 
$p_k \neq 0,0.5$ is an essential assumption for causal 
identification in our setting as will be shown later, as 
an analogue to the aforementioned 
non-Gaussianity in the case of LiNGAM. The model provided by 
Definition~\ref{dfbexsam} and Assumption~\ref{amskew} is called 
a binary EXOR skew acyclic model, or ``{\it BExSAM}'' for short. 

If $f_k$ in Definition~\ref{dfbexsam} depends on $x_h$ $(h<k)$, 
we say that $x_h$ is a ``{\it parent}'' of $x_k$ and $x_k$ is 
a ``{\it child}'' of $x_h$. As is widely noted in causal inference 
studies~\cite{Pearl:2000, Spirtes:2000}, we divide 
$X=\{x_k|x=1,\dots,d\}$ into two classes: $x_k$ having no 
parents (``{\it exogenous variables}'') and $x_k$ 
having some parents (``{\it endogenous variables}'').  
In this study, we further introduce the following 
definition of a particular endogenous variable.
\begin{df}
Endogenous variables having no {\it children} are  called 
``{\it sinks}.''\hfill $\bsquare$
\end{df}
As shown in our later discussion, finding sink endogenous variables 
plays a key role with regard to principles and algorithms for identification 
of a unique causal order and estimation of a BExSAM .

\noindent {\bf Example 1} The following is an example of 
a BExSAM consisting of four binary variables. 
\begin{eqnarray}
x_1 & = & e_1, \nonumber\\
x_2 & = & x_1 \oplus e_2, \nonumber\\
x_3 & = & x_1x_2 \oplus e_3, \nonumber\\
x_4 & = & (x_1 + x_3) \oplus e_4, \nonumber
\vspace{-13mm}
\end{eqnarray}
where the values of $x_1x_2$ and $x_1 + x_3$ 
are given in Table~1. As shown in Fig.~1, $x_1$ and $x_4$ are 
exogenous and sink endogenous variables, respectively. 
\begin{table}[t]\label{truthtable} 
\begin{center} 
\caption{A truth table of $a \oplus b$, $ab$ and $a+b$.}
\begin{tabular}{cc|ccc} 
\hline 
$a$&$b$&$a \oplus b$&$ab$&$a+b$\\
\hline 
0&0&0&0&0\\
0&1&1&0&1\\
1&0&1&0&1\\
1&1&0&1&1\\
\hline 
\end{tabular} 
\end{center} 
\end{table} 
\begin{figure}[t]\label{exampledag}
\begin{center} 
\includegraphics[scale=0.45,clip]{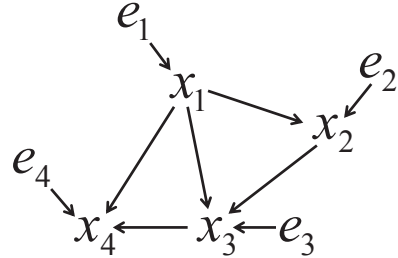}
\vspace{-3mm}
\end{center}
\caption{A DAG structure of the BExSAM in Example 1.}
\end{figure}

\subsection{Characterization}

In this subsection, characteristics of BExSAM associated 
with sink endogenous variables concerning the identification 
of a unique causal order and the estimation of a structural 
model are analyzed. First, we define a notion of 
``{\it selection}'' which specifies the values of some 
variables in $X$. 
\begin{df}
For $k=1,\dots,d$, we denote by $X_k = V_k$ an assignment
$x_1=v_1,\dots,x_{k-1}=v_{k-1},x_{k+1}=v_{k+1},\dots,x_d=v_d$ 
for $X_k:=X\setminus\{x_k\}$ and 
$V_k:=(v_1,\dots,v_{k-1},v_{k+1},\dots,v_d)^T \in \{0, 1\}^{d-1}$.
This assignment is called a ``{\it selection}'' of 
$X_k$ at $V_k$.\hfill $\bsquare$
\end{df}

The following theorem is important for causal ordering 
of the variables in $X$ by using the selection. 
\begin{tm}\label{tm1}
The following conditions are equivalent.
\begin{enumerate}
\item $x_k \in X$ is a {\it sink endogenous variable}.
\item There is a common constant $q_k$ such that $p(x_k=1|X_k=V_k)=q_k$ or $1-q_k$ 
(and therefore, $p(x_k=0|X_k=V_k)=1-q_k$ or $q_k$ equivalently) 
for all selections $X_k=V_k \in \{0, 1\}^{d-1}$.\hfill $\bsquare$
\end{enumerate}
Proof. See Appendix~\ref{appendA}.\hfill $\Box$
\end{tm}

For example, if we are given the two selections $X_4=\{x_1,x_2,x_3\}$ at 
$V_4=(0,0,0)$ and $V'_4=(0,0,1)$ in Example 1,
we have the following conditional probabilities for the sink 
endogenous variable.
\begin{eqnarray}
p(x_4=1|X_4=V_4)&=&p_4,\nonumber\\
p(x_4=1|X_4=V'_4)&=&1-p_4,\nonumber
\end{eqnarray}
Actually, $p(x_4=1|X_4=V_4)=p_4 (=q_4)$ or $1-p_4 (=1-q_4)$ 
holds for any $V_4$ in this case. In contrast, 
if we are provided with selections $X_3=\{x_1,x_2,x_4\}$ at $V_3=(0,0,0)$ 
and $V'_3=(0,0,1)$, then 
\begin{eqnarray}
p(x_3=1|X_3=V_3)&=&\frac{p_3p_4}{p_3p_4+(1-p_3)(1-p_4)},\nonumber\\
p(x_3=1|X_3=V'_3)&=&\frac{p_3(1-p_4)}{p_3(1-p_4)+(1-p_3)p_4}\nonumber
\end{eqnarray}
hold. Because $p_3,p_4 \neq 0.5$ by Assumption~\ref{amskew}, 
these probabilities are not equal, and also their sum is not 
unity. Accordingly, no constant $q_3$ or $1-q_3$ can be 
assigned to both $p(x_3=1|X_3=V_3)$ and $p(x_3=1|X_3=V'_3)$ 
in this case. These results reflect Theorem~\ref{tm1}, 
that we can find a sink endogenous variable in $X$ by 
checking the conditional probability of every variable.

Next, we present an important proposition for estimating 
a structural model of $X$. 
\begin{pr}\label{pr1}
Let $x_k \in X$ be a {\it sink endogenous variable}.
\begin{enumerate}
\item $f_k(x_1,\dots,x_{k-1})=1$ under $X_k=V_k$\\
      $\Leftrightarrow$ $p(x_k=1|X_k=V_k)>p(x_k=0|X_k=V_k)$.
\item $f_k(x_1,\dots,x_{k-1})=0$ under $X_k=V_k$\\
      $\Leftrightarrow$ $p(x_k=1|X_k=V_k)<p(x_k=0|X_k=V_k)$.\hfill $\bsquare$
\end{enumerate}
Proof. See Appendix~\ref{appendB}.\hfill $\Box$
\end{pr}
The function $f_k$ under a selection $X_k=V_k$ is constant 
since all of its arguments are constant. Accordingly, the 
probability distribution of $x_k$ under the selection is 
determined by the constant $f_k$ and the fact that $0< p_k <0.5$ 
in Assumption~\ref{amskew}. In Example~1, under a selection 
$X_4=\{x_1, x_2, x_3\}=V_4=\{1,0,0\}$, $f_4=x_1 + x_3=1$ and 
thus $x_4=1 \oplus e_4 =\bar{e}_4$. This implies that 
$p(x_4=1|X_4=V_4)>p(x_4=0|X_4=V_4)$ since $0< p_4 <0.5$. 
On the other hand, $p(x_4=1|X_4=V_4)>p(x_4=0|X_4=V_4)$ implies that 
$0< p(x_4=0|X_4=V_4) <0.5$. This further implies that $f_4=1$ under 
$X_4=V_4$ since $x_4=f_4 \oplus e_4$ and $0< p_4 <0.5$. 
Proposition~\ref{pr1} indicates a way to identify the part 
of a sink endogenous variable $x_k$ and $X_k=V_k$ in the truth 
table of $f_k$.

\section{Proposed Algorithms}\label{alg_mea}

\subsection{Problem Setting}\label{problem}

First, we define our problem of causal order identification and 
structural model estimation for a BExSAM. 

In our setting, the causal order $x_1,\cdots,x_d$ is unknown in 
advance, but we have a data set $D$ containing a finite number 
of instances $V=(v_{i(1)},\dots,v_{i(d)}) \in \{0,1\}^d$ of variables 
$X=\{x_{i(1)},\cdots,x_{i(d)}\}$ where $i(k)$ labels 
a variable $x_k$ while $k$ is unknown. $i(k)$ is a permutation 
$i:\{1,\dots,d\} \rightarrow \{1,\dots,d\}$ to be identified from 
$D$ in determining the causal order. In addition, Boolean functions 
$f_{i(k)}$ for all $k=1,\dots,d$ in the BExSAM need to be estimated. Note 
that the values of $\{e_{i(1)},\cdots,e_{i(d)}\}$ can be estimated only 
from $D$. $D$ is generated through a process well modeled by a BExSAM 
where the distributions of $\{e_{i(1)},\cdots,e_{i(d)}\}$ are skew and 
jointly independent. Accordingly, if the sample size $n=|D|$ is 
sufficiently larger than $2^d$, then $D$ contains varieties of 
instances $V$ which enables estimation of the conditional probabilities 
under various selections similar to the other constraint based 
approaches~\cite{Spirtes:2000}. 

In summary, we assume that a given data set $D=\{V^{(h)}|h=1,\dots,n\}$ 
is generated in a BExSAM: 
\[x_{i(k)} = f_{i(k)}(x_{i(1)},\dots,x_{i(k-1)}) \oplus e_{i(k)}\qquad (k=1,\dots,d)\]
where $f_{i(1)}$ is a constant in $\{0,1\}$ and 
$f_{i(k)}:\{0,1\}^{k-1} \rightarrow \{0,1\}$ $(k \geq 2)$ is 
deterministic Boolean function. Our problem is to identify 
the permutation $i:\{1,\dots,d\} \rightarrow \{1,\dots,d\}$, 
that is the causal order, and to estimate the functions 
$f_{i(k)}$ ($k=1,\dots,d$) only from $D$.

\subsection{Outline of Proposed Algorithm}\label{outline}
\begin{figure}[t]\label{mainalg}
\begin{center} 
\begin{tabular}{l} 
\hline
input: a binary data set $D$ and its variable list $X$.\\
1. compute a frequency table $FT$ of $D$.\\
2. for $k:=d$ to $1$ do\\
3. \hspace*{2mm} $i(k) := {\bf find\_sink}(FT,X)$.\\ 
4. \hspace*{2mm} $TT_{i(k)} := {\bf find\_truth\_table}(FT,X,i(k))$.\\
5. \hspace*{2mm} remove $x_{i(k)}$ from $X$\\
   \hspace*{6mm} and marginalize $FT$ over $x_{i(k)}$.\\
6. end\\
output: a list $[\{x_{i(k)},TT_{i(k)}\}|k=1,\dots,d]$.\\
\hline
\end{tabular} 
\end{center} 
\caption{Main algorithm.} 
\end{figure}
We propose an approach to solving our problem based on Theorem~\ref{tm1} 
and Proposition~\ref{pr1}\footnote{Code is available from 
http://www.ar.sanken.osaka-u.ac.jp$/$\~\ 
\hspace{-2mm} inazumi/bexsam.html.}. 
Figure~2 shows the outline of our proposed algorithm. Since we 
only need the values of $FT$ rather than $D$ to identify the 
order $x_1,\cdots,x_d$, we compute the values of $FT$ in the 
first stage of the procedure. In the loop from the next step, 
the algorithm seeks a sink endogenous variable $x_{i(k)}$ using 
the function ``find\_sink'' at step 3 and a Boolean function 
$f_{i(k)}$ in the form of a truth table $TT_{i(k)}$ via the function 
``find\_truth\_table'' at step 4. These functions perform 
the identification of a unique causal order and the estimation of 
a {\it BExSAM} entailed by the order. Step 5 reduces the 
search space in the next loop by removing the estimated sink 
endogenous variable $x_{i(k)}$ from $X$ and marginalizing 
$x_{i(k)}$ in $FT$. The entire list of $x_{i(k)}$ and $TT_{i(k)}$ 
in the output represents the causal order of the variables in 
the causal DAG structure and  the {\it BExSAM} reflecting the 
structure. This iterative reduction from the bottom in the causal 
order is similar to the causal ordering of \cite{Mooij:2009}. 
However, their causal structure estimation needs a second sweep 
from the top to the bottom. We should note here that our approach
consisting of the main algorithm, find\_sink and find\_truth\_table 
does not require any parameters to be tuned as shown in the next 
subsection.
\begin{figure}[t]\label{sinkalg}
\begin{center} 
\begin{tabular}{l} 
\hline
input: a frequency table $FT$ and its variable list $X$.\\
1. for $i:=1$ to $|X|$ do\\
2. \hspace*{2mm} $X_i:=X \setminus \{x_i\}$.\\
3. \hspace*{2mm} compute $p_s(x_i=v_i,X_i=V_i)$, $p_s(x_i=v_i)$,\\
   \hspace*{6mm} $p(X_i=V_i)$ for all $v_i \in \{0,1\}$ and\\ 
   \hspace*{6mm} $V_i \in \{0,1\}^{|X|-1}$ from $FT$.\\
4. \hspace*{2mm} compute independence measure $MI_s(x_i,X_i)$.\\
5. end\\
6. select $i$ having the minimum $MI_s(x_i,X_i)$ in $X$.\\
output: $i$.\\
\hline
\end{tabular} 
\end{center} 
\caption{Algorithm for find\_sink.}
\end{figure}
\begin{figure}[h]\label{sortcpmainalg}
\begin{center} 
\includegraphics[scale=0.42,clip]{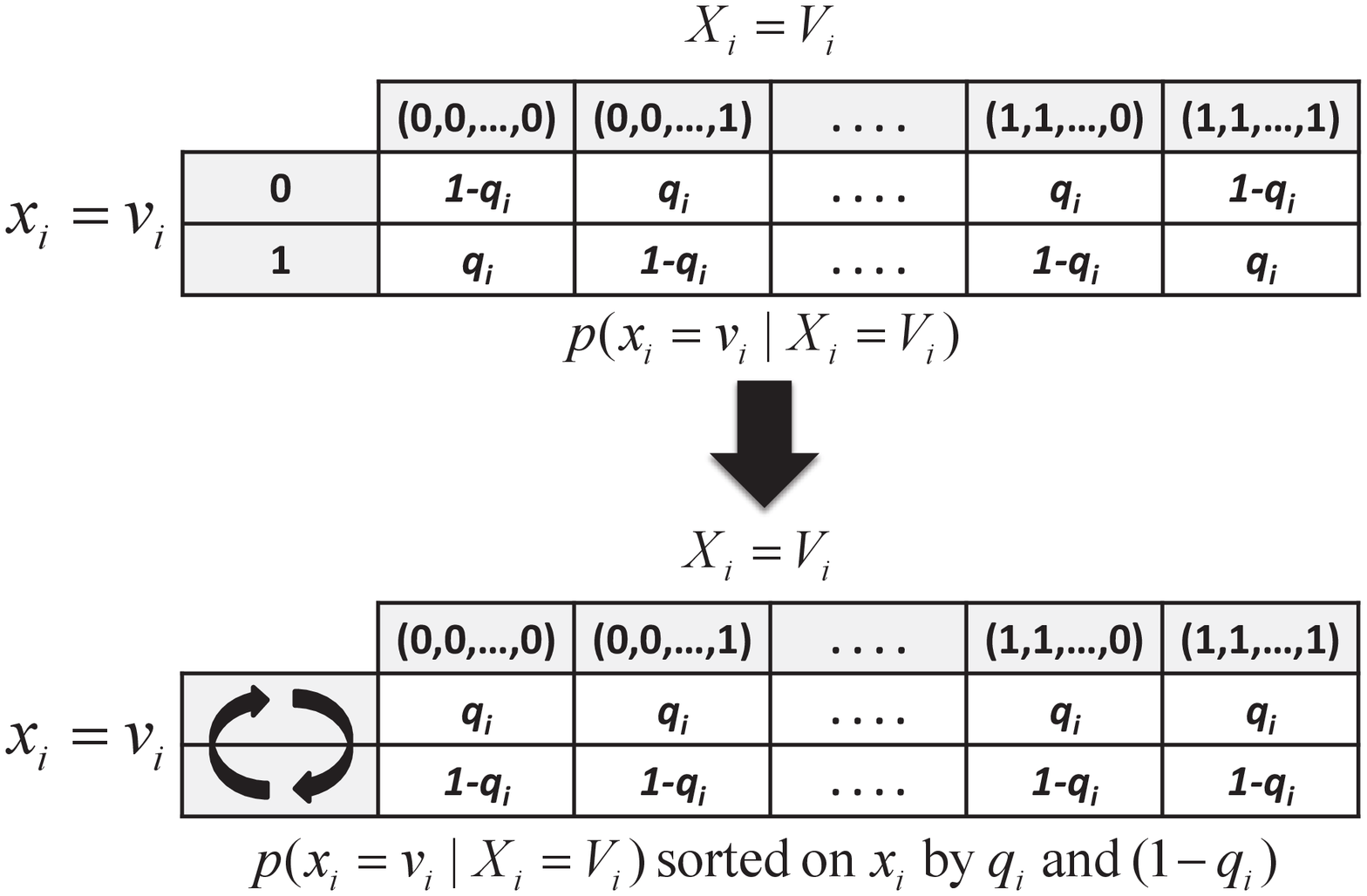}
\vspace{-7.5mm}
\end{center}
\caption{Sort on $x_i$ in a conditional probability table.}
\end{figure}

\subsection{Finding a Unique Causal Order and Functions}
Our algorithm for finding a sink endogenous variable is summarized 
in Fig.~3. In the loop starting from step 1, mutual information 
adapted to our problem: $MI_s(x_i,X_i)$ is computed for each $x_i$ 
from $FT$ as explained below. This represents the dgree to which 
$x_i$ fits condition 2 in Theorem~\ref{tm1}. Finally, $x_i$ 
with the minimum value of $MI_s(x_i,X_i)$, that is the highest 
possibility of being a sink endogenous variable, is selected.
\begin{figure}[t]\label{ttalg}
\begin{center} 
\begin{tabular}{l}
\hline
input: a frequency table $FT$, its variable list $X$\\
\hspace*{10mm} and an index of a {\it sink endogenous variable} $i$.\\
1. $X_i:=X \setminus \{x_i\}$ and $TT_i=\phi$.\\
2. for all $V_i \in \{0,1\}^{|X|-1}$ do\\
3. \hspace*{2mm} compute $p(x_i=v_i|X_i=V_i)$ for all $v_i \in \{0,1\}$\\
   \hspace*{6mm}  from $FT$.\\
4. \hspace*{2mm} If $p(x_i=1|X_i=V_i)>p(x_i=0|X_i=V_i)$, $f_i=1$,\\
   \hspace*{6mm} otherwise $f_i=0$.\\
5. \hspace*{2mm} $TT_i=TT_i + \{f_i\}$.\\
6. end\\
output: $TT_i$.\\
\hline
\end{tabular} 
\end{center} 
\caption{Algorithm for find\_truth\_table.}
\end{figure}
\begin{figure}[h]\label{ttable}
\begin{center} 
\includegraphics[scale=0.5,clip]{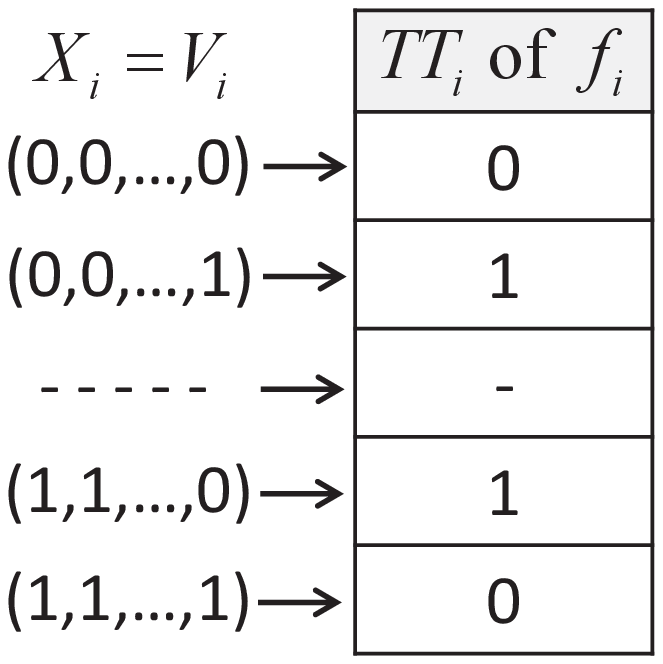}
\vspace{-3mm}
\end{center}
\caption{A truth table of $f_i$.}
\end{figure}

As noted in Theorem~\ref{tm1}, $p(x_i=1|X_i=V_i)$ takes one of the values 
$q_i$ or $1-q_i$ depending on the selections as depicted in the upper 
table in Fig.~4, if and only if $x_i$ is a sink endogenous variable. We 
then obtain the bottom table which represents the independence of 
$x_i$ from $X_i$ after sorting $p(x_i=v_i|X_i=V_i)$ 
in ascending order in every column according to $q_i < 1-q_i$. This 
implies that the independence of $x_i$ from $X_i$ in the table sorted 
on $x_i$ is equivalent to the fact that $x_i$ is a sink endogenous 
variable. Practically, if the frequencies of some $X_i=V_i$ are zero in 
$FT$ because of the incomplete cover of the selections of $X_i=V_i$ in 
the given data set $D$, the probabilities on such $X_i=V_i$ are not 
computable. Thus, we obtain the probabilities on $N_i$ selections of 
$X_i=V_i$ less than $2^{d-1}$. Based on these considerations, we use 
the following mutual information $MI_s(x_i,X_i) \geq 0$ between the 
sorted $x_i$ and $X_i$ to evaluate the dgree to which $x_i$ is a sink 
endogenous variable. 
\begin{eqnarray}
MI_s(x_i,X_i)&=&\frac{2^{d-1}}{N_i}\sum_{v_i,V_i}\biggl\{p_s(x_i=v_i,X_i=V_i)\biggr.\nonumber\\
&& \left. \times \ln\frac{p_s(x_i=v_i,X_i=V_i)}{p_s(x_i=v_i)p(X_i=V_i)}\right\},\nonumber
\end{eqnarray}
where $p_s$ represents a probability for the sorted $x_i$ and the 
summation is taken over the available $N_i$ selections. Because 
$MI_s(x_i,X_i)$ is rewritten as
\begin{eqnarray}
MI_s(x_i,X_i)&=&\frac{2^{d-1}}{N_i}\sum_{v_i,V_i}\biggl\{p_s(x_i=v_i|X_i=V_i)\biggr.\nonumber\\
&& \left. \times p(X_i=V_i)\ln\frac{p_s(x_i=v_i|X_i=V_i)}{p_s(x_i=v_i)}\right\},\nonumber
\end{eqnarray}
it is zero when $p_s(x_i=v_i|X_i=V_i)=p_s(x_i=v_i)$ as in the bottom table 
in Fig.~4. A smaller $MI_s(x_i,X_i)$ represents a higher possibility of 
$x_i$ being a sink endogenous variable. 

Figure~5 outlines our algorithm for estimating every function $f_i$. 
In the loop beginning from step 2, the conditional probability of $x_i$ 
for each selection $X_i=V_i$ is computed at step 3, and the value of 
$f_i$ is estimated by following Proposition~\ref{pr1} at step 4. This is 
further listed in a predefined order on $V_i$ in a truth table 
$TT_i$ at step 5 as depicted in Fig.~6. Similarly to the former algorithm 
of find\_sink, we assign `void' to $f_i$ when the frequency of 
$X_i=V_i$ is zero by the incompleteness of $FT$. The final output holds 
the entire truth table of $f_i$.

\subsection{Computational Complexity}
The largest table used in the above algorithms is the frequency 
table $FT$ which has size $2^d$. Thus, the memory complexity of 
our algorithms is $O(2^d)$. According to the requirement of data 
size, $n \geq 2^d$, as noted in section~\ref{problem}, this is 
also written as $O(n)$. 

The loop involved in the ``find\_sink'' function computes the 
probabilities at most $2^{d-1}$ times, and computes the independence 
measure $MI_s(x_i,X_i)$ by aggregating these $2^{d-1}$ probabilities. 
Thus, this process is $O(2^d) \simeq O(n)$. Since the loop repeats 
$d$ times at most, the time complexity of ``find\_sink'' is 
$O(d2^d) \simeq O(dn)$. The loop involved in ``find\_truth\_table'' 
function computes the conditional probabilities and estimates an 
element in the truth table at most $2^{d-1}$ times. Therefore, 
the time complexity of ``find\_truth\_table'' is $O(2^d) \simeq O(n)$. 
Step 1 of the main algorithm needs $n \geq 2^d$ counts to compute 
$FT$, so is $O(2^d) \simeq O(n)$. The functions ``find\_sink'' 
and ``find\_truth\_table'' which are $O(d2^d) \simeq O(dn)$ and 
$O(2^d) \simeq O(n)$ are repeated $d$ times in the main algorithm. 
Accordingly, the total time complexity of the proposed algorithms is 
$O(d^22^d) \simeq O(d^2n)$. 

This computational complexity is tractable when the number of 
observed variables is moderate as shown in the numerical experiments 
later. This complexity is also favorable compared with 
past work. For example, DirectLiNGAM which also has an iterative 
algorithm structure and is considered to be one of the most efficient 
algorithms has $O(d^3n)$ complexity. 

\section{Experimental Evaluation}

\subsection{Basic Performance for Artificial Data}\label{basicp}
For our numerical experiments, we generated artificial data sets using 
{\it BExSAMs} produced by the following procedure. For every $f_k$ 
($k>1$) in Definition~\ref{dfbexsam}, we randomly choose each $x_i$ 
from the set of potential ancestors $X^{1:k-1}=\{x_i|i=1,\dots,k-1\}$ 
as a parent of $x_k$ with probability $p_a$. Given a set of 
parents $X^{Pa}_k \subseteq X^{1:k-1}$ chosen in this way, 
we set $f_k$ to $0$ or $1$ uniformly at random for all selections 
$X^{Pa}_k=V^{Pa}_k \in \{0,1\}^{|X^{Pa}_k|}$. We do not care about 
the other non-parent variables in $X^{1:k-1}$ when defining $f_k$, 
and obtain a truth table 
$TT_k=\{f_k|X^{1:k-1}=V^{1:k-1} \in \{0,1\}^{d-1}\}$ 
based on $f_k$ for all $V^{Pa}_k$. For $f_1$, we simply form 
the truth table $TT_1=\{f_1\}$ where $f_1$ is a constant 
chosen from $\{0,1\}$ uniformly at random. This random procedure 
generates a generic {\it BExSAM} in the form of a truth table 
$TT=\{TT_k|k=1,\dots,d\}$. 

We obtained our artificial data set $D=\{V^{(h)}|h=1,\dots,n\}$ 
from the generated {\it BExSAM} in the following way. We randomly 
generate $e^{(h)}_k$ ($k=1,\dots,d$ and $h=1,\dots,n$) under 
respective $p_k \in (0,0.5)$ which are common over all $h$ by 
Assumption~\ref{amskew}. For each $h$, we successively derive 
the value of $f_k$ from $k=1$ to $d$ by applying the values 
of $X^{1:k-1}$ to $TT_k$, and compute the value of $x_k$ by 
$f_k \oplus e_k$. Once this tentative data set is obtained, 
we randomly permute the indices $k=1,\dots,d$ of the causal 
ordering to define new variable indices $i(k)$ ($k=1,\dots,d$), 
and obtain the final data set $D$. The series of model generation, 
data generation and application of our approach was repeated 1000 
times for various combinations of the parameters $d$, $n$, $p_a$ 
and $p_{i(k)}$ ($k=1,\dots,d$).

\begin{table*}[t] 
\caption{Performance under various $n$ and $d$ 
(top:$F(A)$, middle:$F(TT)$ and bottom:$CT$ (msec) in each cell).} 
\label{result1} 
\vspace{-3mm}
\begin{center} 
\begin{tabular}{rllllllllll} 
\multicolumn{1}{c}{\bf $n \backslash d$}&\multicolumn{1}{c}{2}&\multicolumn{1}{c}{4}&\multicolumn{1}{c}{6}&\multicolumn{1}{c}{8}&\multicolumn{1}{c}{10}&\multicolumn{1}{c}{12}&\multicolumn{1}{c}{14}&\multicolumn{1}{c}{16}&\multicolumn{1}{c}{18}&\multicolumn{1}{c}{20}\\ 
\hline 
&{\bf 1.000}&0.737&0.658&0.550&0.444&0.386&0.368&0.371&0.395&0.417\\
\cline{2-11} 100&{\bf 0.932}&0.802&0.686&0.563&0.465&0.404&0.359&0.323&0.299&0.274\\
\cline{2-11} &{\bf 0.521}&{\bf 1.14}&{\bf 2.20}&{\bf 4.61}&{\bf 13.7}&{\bf 57.4}&{\bf 233}&1150&6158&32197\\
\hline 
&{\bf 1.000}&0.893&{\bf 0.912}&0.867&0.734&0.571&0.458&0.386&0.351&0.339\\ 
\cline{2-11} 500&{\bf 0.971}&{\bf 0.909}&0.860&0.771&0.619&0.498&0.427&0.386&0.353&0.328\\
\cline{2-11} &{\bf 0.462}&{\bf 1.13}&{\bf 2.19}&{\bf 4.59}&{\bf 13.6}&{\bf 57.6}&{\bf 234}&1155&6192&32321\\
\hline 
&{\bf 1.000}&{\bf 0.917}&{\bf 0.941}&{\bf 0.933}&0.851&0.704&0.559&0.454&0.382&0.346\\ 
\cline{2-11} 1000&{\bf 0.980}&{\bf 0.934}&0.890&0.829&0.715&0.570&0.472&0.411&0.374&0.348\\
\cline{2-11} &{\bf 0.446}&{\bf 1.14}&{\bf 2.18}&{\bf 4.59}&{\bf 13.6}&{\bf 57.6}&{\bf 235}&1156&6206&32338\\
\hline 
&{\bf 1.000}&{\bf 0.962}&{\bf 0.978}&{\bf 0.984}&{\bf 0.970}&{\bf 0.905}&0.783&0.661&0.553&0.464\\ 
\cline{2-11} 5000&{\bf 0.986}&{\bf 0.966}&{\bf 0.946}&{\bf 0.910}&0.856&0.749&0.600&0.495&0.432&0.396\\
\cline{2-11} &{\bf 0.460}&{\bf 1.17}&{\bf 2.25}&{\bf 4.71}&{\bf 14.0}&{\bf 54.2}&{\bf 242}&1174&6195&32280\\
\hline 
&{\bf 1.000}&{\bf 0.971}&{\bf 0.987}&{\bf 0.990}&{\bf 0.984}&{\bf 0.956}&0.860&0.738&0.632&0.534\\ 
\cline{2-11} 10000&{\bf 0.994}&{\bf 0.976}&{\bf 0.959}&{\bf 0.933}&0.888&0.815&0.674&0.544&0.464&0.417\\
\cline{2-11} &{\bf 0.474}&{\bf 1.17}&{\bf 2.25}&{\bf 4.70}&{\bf 14.0}&{\bf 54.0}&{\bf 242}&1166&6202&32234\\
\hline 
\end{tabular} 
\end{center} 
\end{table*} 

Three performance indices were used for the evaluation. Given two binary 
adjacency matrices $A_t$ and $A_e$ representing the parent-child 
relationships between the variables in the generated {\it BExSAM} 
and its estimated {\it BExSAM} respectively, we compute their precision 
and recall as follows.
\[P(A) = |A_t \wedge A_e|/|A_e| \mbox{ and } R(A) = |A_t \wedge A_e|/|A_t|,\]
where $\wedge$ is an element-wise AND operation and $|\cdot|$ is the 
number of non-zero elements in a matrix. We then obtain their resultant 
$F$-measure as the first performance index.
\[F(A) = 2\frac{P(A) \cdot R(A)}{P(A)+R(A)}.\]
This represents the performance of the causal ordering. Similarly, we 
compute an $F$-measure $F(TT)$ between the true truth table $TT_t$ and 
the estimated truth table $TT_e$ as the second performance index, where 
$f_i$ having the `void' values in both $TT_e$ and its corresponding $f_i$ 
in $TT_t$ were skipped in the element-wise AND operation of $\wedge$. 
This indicates the performance of the model estimation. The third index 
is simply the total computational time $CT$ (msec) of our algorithm explained 
in section~\ref{alg_mea}. These indices are averaged over the 1000 trials. 

In the first experiment, every combination of 
$d=2,4,6,8,10,12,14,16,18,20$ and $n=100,500,1000,$ $5000,10000$ 
was evaluated with $p_a=0.5$ and $p_{i(k)}$ 
defined uniformly at random over $(0,0.5)$ for $k=1,\dots,d$. 
These choices of $p_{i(k)}$ ensure the skew and non-deterministic 
distribution of $e_{i(k)}$ as required by Assumption~\ref{amskew}. 
$p_a$ reflects the density of the variable couplings in the generated 
{\it BExSAM}. Table~\ref{result1} summarizes the performance of 
our approach. Values of $F(A)$ and $F(TT)$ greater than $0.9$ are typed 
in boldface, and values of $CT$ less than $1000$ msec are also written 
in boldface. We observe that $F(A)$ can be less than $0.9$ even 
with $n > 2^d$ for a small $d$. This is because the statistical accuracy 
of $MI_s(x_i,X_i)$ is not very high according to the summation over 
the small number $N_i$($\simeq 2^d$) of $V_i$. The accuracy of 
the causal ordering is affected by erroneous values for $MI_s(x_i,X_i)$.
On the other hand, $F(A)$ is greater than $0.9$ under $n > 2^d$ when 
$d$ is large because of the higher accuracy of $MI_s(x_i,X_i)$. 
We further observe that $F(TT)$ is more sensitive to shortages in 
the data than $F(A)$. This is because the estimation accuracy of 
$p(x_i=v_i|X_i=V_i)$ required to derive $TT$ is directly affected 
by the frequency of the individual $X_i=V_i$ in the data. The 
estimation accuracy is strongly reduced by the smaller frequency 
for smaller $n$. These results indicate that our causal ordering 
approach and model estimation approach work properly for up to $12$ 
variables and up to 8 or 10 variables, respectively, with a 
dataset of several thousands samples. We note that in particular 
$CT$ increases with $d$ but not with $n$. This is 
consistent with the aforementioned complexity analysis and the 
fact that $n$ affects only the computation of $FT$ in the initial 
stage of our algorithm. The results show that our algorithm 
completes the causal ordering and the model estimation within 
a second up to $d=14$ even for large amounts of data.

\begin{figure}
\begin{center}
\includegraphics[scale=0.65,clip]{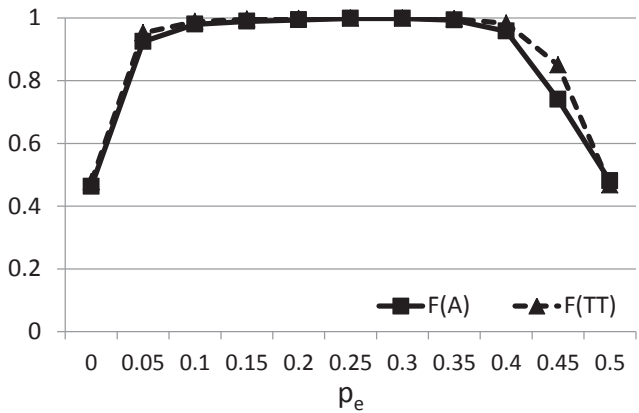}
\vspace{-5mm}
\caption{Dependency of $F(A)$ and $F(TT)$ on $p_e$.}\label{deppe}
\end{center}
\end{figure}

In the second experiment, we gave an identical value 
$p_e \in (0,0.5)$ to $p_{i(k)}$ for all $k=1,\dots,d$, 
and evaluated $F(A)$ and $F(TT)$ for $d=4$, $n=1000$ and 
$p_a=0.5$. Figure~\ref{deppe} depicts the resultant dependency 
of $F(A)$ and $F(TT)$ on various $p_e$. If $p_e$ is close to 
$0$, our approach fails to accurately estimate 
the conditional probabilities and thus its accuracy is degraded. 
If $p_e$ is closed to 0.5, again the accuracy of our approach 
is lost, because it relies heavily on Theorem~\ref{tm1} 
and Proposition~\ref{pr1} which require $p_{i(k)} \neq 0.5$. 
Through some extra experiments, we confirmed that $F(A)$ and $F(TT)$ 
do not show strong dependency on $p_a$, the causal 
density of the {\it BExSAM}. 

\begin{table}[t] 
\caption{Comparison with other algorithms.} 
\label{result2} 
\vspace{-2mm}
\begin{center} 
\begin{tabular}{lrrr}
\hline 
\multicolumn{4}{l}{Our Algorithm}\\
\hline 
\multicolumn{1}{c}{\bf true $\backslash$ est. }&\multicolumn{1}{c}{directed}&\multicolumn{1}{c}{no edge}&\multicolumn{1}{c}{undirected}\\ 
\hline 
directed &{\it 55}&5&0\\
\hline 
no edge &8&{\it 172}&0\\ 
\hline 
\hline 
\multicolumn{4}{l}{PC Algorithm}\\
\hline 
\multicolumn{1}{c}{\bf true $\backslash$ est. }&\multicolumn{1}{c}{directed}&\multicolumn{1}{c}{no edge}&\multicolumn{1}{c}{undirected}\\ 
\hline 
directed &{\it 28}&18&14\\
\hline 
no edge &6&{\it 158}&16\\ 
\hline 
\hline 
\multicolumn{4}{l}{CPC Algorithm}\\
\hline 
\multicolumn{1}{c}{\bf true $\backslash$ est. }&\multicolumn{1}{c}{directed}&\multicolumn{1}{c}{no edge}&\multicolumn{1}{c}{undirected}\\ 
\hline 
directed &{\it 27}&12&21\\
\hline 
no edge &0&{\it 157}&23\\ 
\hline 
\hline 
\multicolumn{4}{l}{GES Algorithm}\\
\hline 
\multicolumn{1}{c}{\bf true $\backslash$ est. }&\multicolumn{1}{c}{directed}&\multicolumn{1}{c}{no edge}&\multicolumn{1}{c}{undirected}\\ 
\hline 
directed &{\it 24}&17&19\\
\hline 
no edge &0&{\it 161}&19\\ 
\hline 
\end{tabular} 
\end{center} 
\end{table} 

\subsection{Comparison with Other Algorithms}
Our algorithm, the PC algorithm~\cite{Spirtes:2000}, 
the CPC algorithm~\cite{Ramsey:2006} and the 
GES algorithm~\cite{Chickering:2002} were 
compared by applying them to data generated 
by the following artificial {\it BExSAM} forming 
a Y-structure~\cite{Mani:2006}.
\begin{eqnarray}
x_1&=&e_1\nonumber\\
x_2&=&e_2\nonumber\\
x_3&=&x_1 x_2 \oplus e_3\nonumber\\
x_4&=&x_3 \oplus e_4\nonumber
\end{eqnarray}
Each $p_{i(k)}$ was given similarly to the first 
experiment in the previous subsection. The significance levels 
$\alpha$ in both PC and CPC were set at $0.05$. 
Table~\ref{result2} shows the frequencies of estimated relationships 
between variables over their true relationships for 20 trials. 
The columns and rows represent estimated relationships and true 
relationships, respectively. Because of the Y-structure among the 
four variables, the number of true directed edges is $3 \times 20=60$ 
in total while the number of true non-edges is 
$({{4} \atopwithdelims<> {2}} -3) \times 20=180$ by 
double counting the two missing directed edges between two variables 
for a non-edge. This counting method gives double penalties to 
an incorrect estimation of an edge direction which often comes from 
causal ordering failures affecting the global structure 
estimation. The {\it italics} show the numbers of correct 
estimations. Note that this Y structure is a typical example 
which enables a valid estimation of the PC algorithm. However, 
our approach based on the skewness of the binary data 
distribution provides better accuracy. Similar advantageous 
results of our approach was obtained for the case where 
$x_3=(x_1 +  x_2) \oplus e_3$. The results of CPC and GES are 
similar to PC, since CPC and GES do not have any significant 
advantages over PC at identifying Y-structures.

\subsection{Example Applications to Real-World Data}
Our approach has been applied to two real-world data sets.
One is on leukemia deaths and survivals ($LE=1/0$) 
in children in southern Utah who have high/low exposure to 
radiation ($EX=1/0$) from the fallout of nuclear tests 
in Nevada~\cite{Finkelstein:1990,Pearl:2000}. 
As this contains only two binary variables, conventional 
constraints/score-based approaches cannot estimate any 
unique causal structure. In contrast, our approach found 
a causal order $EX \rightarrow LE$ consistent 
with our intuition. 

\begin{figure}
\begin{center}
\begin{picture}(100,50)(-10,0)
\put(0,40){\makebox(20,10){IQ}}
\put(0,0){\makebox(20,10){PE}}
\put(80,0){\makebox(20,10){CP}}
\put(10,40){\vector(0,-1){30}}
\put(20,5){\vector(1,0){60}}
\put(20,40){\vector(2,-1){60}}
\end{picture}
\caption{Discovered causal structure among $CP$, $PE$ and $IQ$.}\label{colledge}
\end{center}
\end{figure}
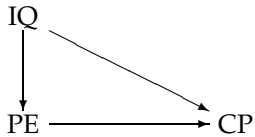

Another data set is for college plans of 10318 Wisconsin 
high school seniors~\cite{Sewell:1968, Spirtes:2000}. 
While the original study aimed to find a feasible causal 
structure among five variables constituting a Y structure, 
we focus on the causality between three variables: yes/no 
college plans ($CP=0/1$), low/high parental encouragement 
($PE=0/1$) and least to highest intelligence quotient 
($IQ=0,\dots,3$). Conventional constraints-based approaches 
are known to give multiple candidate causal structures 
in an equivalence class for the three variables. We selected 
2543 male seniors ($SEX=male$) having a higher socioeconomic 
status ($SES \geq 2$ where $SES=0,\dots,3$) to retain 
individuals having similar social background while maintaining 
an appropriate sample size. We further transformed $IQ$ to a binary 
variable using a threshold value between $1$ and $2$, which gives 
a data set containing 1035 and 1508 seniors having 
$IQ=0$ (lower IQ) and $IQ=1$ (higher IQ), respectively. The 
application of our approach to this data set produced the unique 
causal structure depicted in Fig.~\ref{colledge}. This states 
that the intelligence quotient of a senior affects both his 
parental encouragement and his college plan, and that the parental 
encouragement further influences the college plan. This is 
consistent with our intuition and the structure estimated by the 
PC algorithm from the original five variables constituting a Y 
structure. 

\section{Discussion}
When we apply our approach to a data set, we assume that the 
data generation process approximately follows a {\it BExSAM}. 
A crucial property required in a {\it BExSAM} is the skew 
distribution of every external noise. Because the noise is 
not directly observable, a measure to check this property 
using the given data set is desirable. The following 
lemma can be used for this purpose. 
\begin{lm}\label{lm2}
Assuming that a data set $D$ is generated by a {\it BExSAM}, 
if a variable $x_k \in X$ has a distribution 
$p(x_k=1) \neq 0.5$, then the following condition hold.
\[\qquad \qquad \quad \; p_k < 0.5 \mbox{ and } p(f_k=1) \neq 0.5. \quad \qquad \qquad \bsquare\]
Proof. See Appendix~\ref{appendC}.\hfill $\Box$
\end{lm}
We simply check whether the frequency of $x_k=1$ is different from 
0.5 for every observed variable in $X$. If it is, then the skewness of 
their noise distribution is ensured. Another strong assumption 
of a {\it BExSAM} is the interventions of external binary noises 
via EXOR operations. However, Lemma~\ref{lm2} also suggests the 
applicability of the {\it BExSAM} to generic Boolean interventions 
of the noises. For example, if a given data is generated by 
$x_k=f_k + e_k$, then we have $p(x_k=1)=p_k+p(f_k=1)-p_kp(f_k=1)$. 
On the other hand, a {\it BExSAM} $x_k=f_k \oplus e_k$ provides 
$p(x_k=1)=p_k+p(f_k=1)-2p_kp(f_k=1)$. Accordingly, these two 
models show very similar distributions of the observed variables, 
if $p_k, p(f_k=1) \ll 0.5$. These conditions can be checked by 
the insights of Lemma~\ref{lm2}.

As mentioned in subsection~\ref{problem}, our algorithm 
requires a complete data set $D$ in principle, whixh is similar 
to other constraint based approaches~\cite{Spirtes:2000}. 
Therefore, to analyze a data set $D$ in which a large portion 
is incomplete, we need to estimate the missing data in $D$ 
by introducing some data completion techniques such 
as~\cite{Bernaards:2007}. Another associated issue is the 
applicability to many variables, since the low error rates 
are ensured only for up to 12 variables with several thousands 
of samples. A promising way to overcome this issue may be to 
combine our approach with other constraint-based approaches 
such as the PC algorithm as discussed in~\cite{Zhang:2009}.
The extensions of our approach toward these issues are topics 
for future studies.

\section{Conclusion}
In this paper, we presented a novel binary structural model 
involving exclusive-or noise and proposed an efficient new 
approach to deriving an identifiable causal structure 
governing a given binary data set based on the skewness of the 
distributions of external noises. The approach has low 
computational complexity and does not require any tunable 
parameters. The experimental evaluation shows promising 
performance for both artificial and real world data sets. 

This study provides an extension of the non-Gaussianity-based 
causal inference for continuous variables to 
causal inference for discrete variables, and suggests 
a new perspective on more generic causal inference.

\appendices
\section{Proof of Theorem~\ref{tm1}}\label{appendA}
{\bf (1 $\Rightarrow$ 2)}\\
Let the value of $x_k$ be $v_k \in \{0,1\}$. Under 
a selection $X_k=V_k$, $f_k$ is a constant in 
$\{0,1\}$. Accordingly, the following 
relation holds by Definition~\ref{dfbexsam}.
\[x_k=v_k \Leftrightarrow v_k=f_k \oplus e_k \Leftrightarrow e_k=v_k \oplus f_k.\]
Because $x_k$ is a {\it sink endogenous variable}, 
$e_k$ and $X_k$ are mutually independent by 
Definition~\ref{dfbexsam}. Under this fact 
and Assumption~\ref{amskew}, 
\begin{eqnarray}
&&p(x_k=v_k|X_k=V_k)=\nonumber\\
&&p(e_k=v_k \oplus f_k)=\left\{
\begin{array}{rl}
p_k, & \mbox{ for } v_k \oplus f_k=1\\
1-p_k, & \mbox{ for } v_k \oplus f_k=0
\end{array}\right.\nonumber
\end{eqnarray}
By letting $q_k=p_k$ or $q_k=1-p_k$, 1 $\Rightarrow$ 2 holds.\\
{\bf (1 $\Leftarrow$ 2)}\\
Assume that $x_k$ is not a {\it sink endogenous variable}. 
Let $X_k$ be partitioned into $X^l_k$ and $X^u_k$ where 
$X^l_k$ is a set of all descendants of $x_k$ in a {\it BExSAM}, 
and $X^u_k$ is the complement of $X^l_k$ in $X_k$. 
Then, the following holds.
\begin{eqnarray}
&&\hspace*{-7mm} p(x_k=v_k|X_k=V_k)=p(x_k=v_k|X^u_k=V^u_k,X^l_k=V^l_k)\nonumber\\
&&\hspace*{-7mm} \mbox{$=\frac{p(X^l_k=V^l_k|x_k=v_k,X^u_k=V^u_k)p(x_k=v_k|X^u_k=V^u_k)}{\sum_{v' \in \{v_k,\bar{v}_k\}}p(X^l_k=V^l_k|x_k=v',X^u_k=V^u_k)p(x_k=v'|X^u_k=V^u_k)}$}, \; (t1)\nonumber
\end{eqnarray}
where $\bar{v}_k=v_k \oplus 1$. Furthermore, let 
$X_j=X \setminus \{x_j\}$ for $x_j \in X^l_k$ be partitioned 
into $X^l_j$ and $X^u_j$ similarly to $X^l_k$ and 
$X^u_k$ for $x_k$. By Definition~\ref{dfbexsam}, 
each $f_j$ for $x_j \in X^l_k$ is given by $X^u_j=V^u_j$, 
and thus $x_j=f_j(X^u_j=V^u_j) \oplus e_j$ for all $x_j \in X^l_k$. 
Accordingly, $X^l_k=V^l_k$ is equivalent to 
$e_j=v_j \oplus f_j(X^u_j=V^u_j)$ for all $x_j \in X^l_k$. 
We rewrite the r.h.s.: $v_j \oplus f_j(X^u_j=V^u_j)$ 
as $l_j(v_k)$, since $x_k$ is an ancestor of $x_j$ 
($x_k\in X^u_j$) and the values of all 
variables except $x_k=v_k$ are constant under 
the selection $X_k=V_k$. Because every $e_j$ is 
independent of its upper variables, 
\[p(X^l_k=V^l_k|x_k=v_k,X^u_k=V^u_k)=\prod_{x_j \in X^l_k} p(e_j=l_j(v_k)).\]
$x_k$ has at least one child $x_h \in X^l_k$ 
where $l_h(1) \neq l_h(0)$ for some selection $X_k=V_k$
from the assumption that $x_k$ is not a {\it sink endogenous variable}. 
Accordingly, 
\begin{eqnarray}
&&\hspace*{-7mm} p(X^l_k=V^l_k|x_k=v_k,X^u_k=V^u_k) = \nonumber\\
&&\hspace*{-7mm} {\footnotesize \left\{
\begin{array}{rl}
p_h\prod_{x_j \in X^l_k, x_j \neq x_h} p(e_j=l_j(v_k)), \; for \; l_h(v_k)=1\\
(1-p_h)\prod_{x_j \in X^l_k, x_j \neq x_h} p(e_j=l_j(v_k)), \; for \; l_h(v_k)=0
\end{array}\right.}.\; (t2)\nonumber
\end{eqnarray}
On the other hand, since $e_k$ is independent of $X^u_k$ and 
$x_k=v_k \Leftrightarrow e_k=v_k \oplus f_k$,
\begin{eqnarray}
&&\hspace*{-7mm} p(x_k=v_k|X^u_k=V^u_k)=p(e_k=v_k \oplus f_k)\nonumber\\
&&\hspace*{11mm} = \left\{
\begin{array}{rl}
p_k, & \; for \; v_k \oplus f_k=1\\
1-p_k, & \; for \; v_k \oplus f_k=0
\end{array}\right..\qquad \quad (t3)\nonumber
\end{eqnarray}
By substituting Eqs.$(t2)$ and $(t3)$ into Eq.$(t1)$, we obtain the following four cases.
\begin{eqnarray}
&&\hspace*{-7mm}p(x_k=v_k|X_k=V_k)=\nonumber\\
&&\hspace*{-7mm}\mbox{{\large $\frac{\alpha p_k p_h}{\alpha p_k p_h + \beta (1-p_k)(1-p_h)} \;$} $(t4)$, $\:$ {\large $\frac{\alpha p_k (1-p_h)}{\alpha p_k(1-p_h) + \beta (1-p_k)p_h} \;$} $(t5)$},\nonumber\\
&&\hspace*{-7mm}\mbox{{\large $\frac{\alpha (1-p_k)p_h}{\alpha (1-p_k)p_h + \beta p_k(1-p_h)} \;$} $(t6)$, $\:$ {\large $\frac{\alpha (1-p_k)(1-p_h)}{\alpha (1-p_k)(1-p_h) + \beta p_kp_h} \;$} $(t7)$},\nonumber
\end{eqnarray}
where $\alpha=\prod_{x_j \in X^l_k, x_j \neq x_h} p(e_j=l_j(v_k))$ 
and $\beta=\prod_{x_j \in X^l_k, x_j \neq x_h} p(e_j=l_j(\bar{v}_k))$.
$\alpha$ and $\beta$ are nonzero from Assumption~\ref{amskew}, 
and each of $(t4) = (t5)$, $(t4) = (t6)$, $(t5) = (t7)$ 
and $(t6) = (t7)$ for any $\alpha$ and $\beta$, that is 
any $X^l_k=V^l_k$ excluding $x_h$, implies that $p_k=1/2$ or 
$p_h = 1/2$ respectively. Because neither $p_k \neq 1/2$ 
nor $p_h \neq 1/2$ is allowed by Assumption~\ref{amskew}, 
this implies that all conditions $(t4) \neq (t5)$, $(t4) \neq (t6)$, 
$(t5) \neq (t7)$ and $(t6) \neq (t7)$ hold simultaneously 
for some $X^l_k=V^l_k$ excluding $x_h$. If we assume that 
$(t4)=(t7)$ and $(t5)=(t6)$ simultaneously for such 
$X^l_k=V^l_k$ excluding $x_h$, then 
$p_k^2p_h^2=(1-p_k)^2(1-p_h)^2$ and $p_k^2(1-p_h)^2=(1-p_k)^2p_h^2$, 
and so $p_k=1/2$ and $p_h=1/2$. Accordingly, 
$p_k \neq 1/2$ and $p_h \neq 1/2$ from Assumption~\ref{amskew} 
imply that one of $(t4)=(t7)$ and $(t5)=(t6)$ do not hold 
for $X^l_k=V^l_k$ excluding $x_h$. This result shows 
that $p(x_k=v_k|X_k=V_k)$ takes more than two values 
for some given selection $X_k=V_k$ if $x_k$ is not a 
{\it sink endogenous variable}. By taking the 
contrapositive, we obtain 1 $\Leftarrow$ 2.\hfill $\Box$

\section{Proof of Proposition~\ref{pr1}}\label{appendB}
\begin{itemize}
\item[{\rm (i)}] Since $f_k$ is binary, $f_k$ only takes the values $1$ 
or $0$ under any selection $X_k=V_k$. This and $x_k=f_k \oplus e_k$ 
deduce the relations $x_k=\bar{e}_k$ or $x_k=e_k$, respectively. 
Accordingly, one of $0<p(x_k=0|X_k=V_k)<0.5$ and 
$0<p(x_k=1|X_k=V_k)<0.5$ holds because $0<p_k<0.5$. This implies 
that $p(x_k=1|X_k=V_k) \neq p(x_k=0|X_k=V_k)$.
\item[{\rm (ii)}] If $f_k=1$, then $x_k=\bar{e}_k$ is deduced from 
$x_k=f_k \oplus e_k$. This implies that $0<p(x_k=0|X_k=V_k)<0.5$ by 
$0<p_k<0.5$ and thus $p(x_k=1|X_k=V_k) > p(x_k=0|X_k=V_k)$.
\item[{\rm (iii)}] If $f_k=0$, then $x_k=e_k$ is deduced from 
$x_k=f_k \oplus e_k$. 
This implies that $0<p(x_k=1|X_k=V_k)<0.5$ by $0<p_k<0.5$ and thus
$p(x_k=1|X_k=V_k) < p(x_k=0|X_k=V_k)$.
\end{itemize}
From {\rm (ii)}, {\rm (iii)} and thier contrapositives with {\rm (i)}, 
the proposition is proved.\hfill $\Box$

\section{Proof of Lemma~\ref{lm2}}\label{appendC}
Without loss of generality, $p(e_k=1)(=p_k)$, $p(e_k=0)(=1-p_k)$, $p(f_k=1)$, 
$p(f_k=0)$ with the definition $p(e_k=1) \leq p(e_k=0)$ are represented as
\[p(e_k=1)=\frac{1+\epsilon_k}{2}, \; p(e_k=0)=\frac{1-\epsilon_k}{2} \quad (-1 \leq \epsilon_k \leq 0),\]
\[p(f_k=1)=\frac{1+\xi_k}{2}, \; p(f_k=0)=\frac{1-\xi_k}{2} \quad (-1 \leq \xi_k \leq 1).\]
Because $e_k$ and $f_k$ are mutually independent, 
and have the relation $x_k=f_k \oplus e_k$, 
\begin{eqnarray}
p(x_k=1)&=&p(e_k=1)p(f_k=0)+p(e_k=0)p(f_k=1)\nonumber\\
&=&\frac{(1+\epsilon_k)(1-\xi_k)}{4}+\frac{(1-\epsilon_k)(1+\xi_k)}{4}\nonumber\\
&=&\frac{1-\epsilon_k \xi_k}{2},\nonumber\\
p(x_k=0)&=&p(e_k=1)p(f_k=1)+p(e_k=0)p(f_k=0)\nonumber\\
&=&\frac{(1+\epsilon_k)(1+\xi_k)}{4}+\frac{(1-\epsilon_k)(1-\xi_k)}{4}\nonumber\\
&=&\frac{1+\epsilon_k \xi_k}{2}.\nonumber
\end{eqnarray}
\[\hspace*{-35mm}\Rightarrow p(x_k=0)-p(x_k=1)=\epsilon_k \xi_k.\]
Accordingly, if $p(x_k=1) \neq p(x_k=0)$ then $\epsilon_k \xi_k \neq 0$. 
This implies that $\epsilon_k < 0$ and $\xi_k \neq 0$. Therefore, 
if $p(x_k=1) \neq 0.5$ then $p(e_k=1)=p_k < 0.5$ and 
$p(f_k=1) \neq 0.5$.\hfill $\Box$

\ifCLASSOPTIONcompsoc
  \section*{Acknowledgments}
\else
  \section*{Acknowledgment}
\fi
This work was partially supported by JST, ERATO, Minato Discrete 
Structure Manipulation System Project and JSPS Grant-in-Aid for 
Scientific Research(B) \#22300054. The authors would like to thank 
Dr. Tsuyoshi Ueno, a research fellow of the JST ERATO project, for 
his valuable technical comments.

\ifCLASSOPTIONcaptionsoff
  \newpage
\fi

\end{document}